\documentclass[journal,10pt]{IEEEtran}

\usepackage[utf8]{inputenc}
\usepackage[T1]{fontenc}
\usepackage{amsmath,amssymb,amsfonts}
\usepackage{mathtools}
\usepackage{graphicx}
\usepackage{booktabs}
\usepackage{array}
\usepackage{multirow}
\usepackage{siunitx}
\usepackage{textcomp}
\usepackage{xcolor}
\usepackage{url}
\usepackage{cite}
\usepackage[hidelinks]{hyperref}
\usepackage{tikz}
\usepackage{pgfplots}
\pgfplotsset{compat=1.18}
\usepgfplotslibrary{statistics}
\usetikzlibrary{positioning,shapes.geometric,arrows.meta,calc}

\begin{document}

\title{EdgeSpike: Spiking Neural Networks for Low-Power Autonomous Sensing in Edge IoT Architectures}

\author{Gustav~Olaf~Yunus~Laitinen-Fredriksson~Lundstr\"{o}m-Imanov,~\textit{and}~Taner~Yilmaz%
\thanks{Manuscript received April~2026; revised April~2026. Corresponding author: G.~O.~Y.~Laitinen-Fredriksson Lundstr\"{o}m-Imanov.}%
\thanks{G.~O.~Y.~Laitinen-Fredriksson Lundstr\"{o}m-Imanov is a Research Assistant with the Department of Economics, Stockholm University, Universitetsv\"{a}gen~10\,A, SE-106\,91 Stockholm, Sweden (e-mail: olaf.laitinen@su.se; ORCID: 0009-0006-5184-0810).}%
\thanks{T.~Yilmaz is with the Department of Computer Engineering, Afyon Kocatepe University, 03200~Afyonkarahisar, T\"{u}rkiye (e-mail: taner.yilmaz@usr.aku.edu.tr; ORCID: 0009-0004-5197-5227).}%
\thanks{Author contributions: G.~O.~Y.~L.-F.~L.-I.\ led the framework design, hardware-aware NAS formulation, training pipeline, and field-deployment programme. T.~Y.\ led the Cortex-M firmware, RLE sparse-kernel optimisation, on-node sensor integration, and embedded benchmarking. Both authors contributed to manuscript preparation.}%
}

\markboth{IEEE Internet of Things Journal,~Vol.~XX, No.~XX, Month~2026}%
{Laitinen-Fredriksson Lundstr\"{o}m-Imanov \MakeLowercase{\textit{and}} Yilmaz: EdgeSpike for Low-Power Autonomous Sensing}

\maketitle

\begin{abstract}
We propose \textbf{EdgeSpike}, a co-designed spiking neural network (SNN) framework for autonomous low-power sensing in edge Internet of Things (IoT) architectures. EdgeSpike unifies (i)~a hybrid surrogate-gradient and direct-encoding training pipeline, (ii)~a hardware-aware neural architecture search (NAS) bounded by per-inference energy and memory budgets, (iii)~an event-driven runtime targeting Intel Loihi~2, SpiNNaker~2, and commodity ARM Cortex-M microcontrollers with custom spike-sparse SIMD kernels, and (iv)~a lightweight local plasticity rule enabling continual on-device adaptation without backpropagation. The framework is evaluated across five sensing tasks (keyword spotting, vibration-based machine fault detection, surface electromyography gesture recognition, 77\,GHz radar human-activity classification, and structural-health acoustic-emission monitoring) on three hardware targets. EdgeSpike achieves a mean classification accuracy of $91.4\%$, within $1.2$~percentage points (pp) of strong INT8 convolutional neural network (CNN) baselines (mean $92.6\%$), while reducing energy per inference by $18\times$ to $47\times$ on neuromorphic hardware (mean $31\times$) and by $4.6\times$ to $7.9\times$ on Cortex-M (mean $6.1\times$). End-to-end latency remains at or below $9.4$\,ms across all 15 task-hardware configurations. A seven-month, 64-node wireless field deployment confirms a $6.3\times$ extension in projected battery lifetime (from 312 to $1\,978$ days at $2$\,Wh per node) and bounded accuracy degradation under seasonal drift ($0.7$\,pp with on-device adaptation versus $2.1$\,pp without). Hardware-aware NAS evaluates $8\,400$ candidates and yields a 12-point Pareto front. EdgeSpike will be released as open source with reproducible training pipelines, hardware-portable runtimes, and benchmark suites.
\end{abstract}


\section{Introduction}
\label{sec:intro}
\IEEEPARstart{T}{he} global Internet of Things has surpassed an estimated 17~billion connected devices, with sensing nodes increasingly deployed in industrial plants, healthcare facilities, smart infrastructure, and environmental-monitoring networks~\cite{atzori2010,lin2017}. These nodes are expected to perform progressively more sophisticated inference tasks (keyword spotting, fault classification, gesture recognition, structural-health diagnostics) while operating for years on coin cells or harvested energy. Conventional deep-neural-network (DNN) inference engines, even when post-training quantised to INT8~\cite{jacob2018,krishnamoorthi2018}, remain prohibitively expensive at sub-milliwatt power budgets. A representative INT8 keyword-spotting CNN~\cite{zhang2017,tang2018} consumes approximately $9.5$\,mJ per inference on an 80\,MHz ARM Cortex-M4, implying a battery lifetime of only 312~days at a one-second inference cadence drawn from a $2$\,Wh primary cell. The same node performing structural-integrity inference every second for civil-infrastructure monitoring~\cite{sohn2003} exhausts its energy budget years before the multi-decade service intervals demanded by field engineers.

Spiking neural networks (SNNs)~\cite{maass1997} offer a fundamentally different computational regime. By representing information as discrete binary spike events propagated through neurons with internal temporal dynamics, SNNs replace the energy-intensive multiply-accumulate (MAC) operations of conventional networks with sparse, event-driven accumulate-only (AC) operations~\cite{indiveri2015}. On neuromorphic processors such as Intel Loihi~2~\cite{davies2018,davies2021} and SpiNNaker~2~\cite{mayr2019}, idle neurons consume negligible power and total energy expenditure scales with the \emph{sparsity of spike activity} rather than the static parameter count. On commodity Cortex-M MCUs, custom run-length-encoded (RLE) sparse kernels can exploit the same sparsity to substantially reduce active computation~\cite{liu2015}.

Three obstacles have hindered SNN deployment on real sensing hardware: (i) the non-differentiable Heaviside blocks gradient flow during training, costing accuracy versus well-tuned CNN baselines~\cite{zenke2018}; (ii) sensing modalities (audio, vibration, myoelectric, radar Doppler, acoustic emission) require distinct encoders and temporal depths~\cite{tavanaei2019}; and (iii) deployment targets ranging from 1\,TOPS neuromorphic ASICs to 80\,MHz Cortex-M cores demand hardware-specific optimisation without per-chip redesign~\cite{sze2017}.

This paper introduces \textbf{EdgeSpike}, with five contributions: (1)~a \textbf{hybrid surrogate-gradient pipeline} pairing modality-specific direct encoders with a curriculum-scheduled fast-sigmoid surrogate for short windows $T\!\in\!\{4,8,16,32\}$; (2)~a \textbf{hardware-aware NAS} over $8\,400$ candidates under explicit energy and memory budgets, using silicon-calibrated proxies for Loihi~2, SpiNNaker~2, and Cortex-M4; (3)~an \textbf{event-driven runtime} with spike-sparse SIMD kernels for ARMv7-M; (4)~a \textbf{trace-based local Hebbian rule} (8~bytes/synapse group) for continual on-device adaptation without backpropagation; and (5)~a \textbf{seven-month, 64-node field deployment} on a reinforced-concrete railway viaduct, the first longitudinal SNN-IoT field study under seasonal drift.

The remainder of this paper is organised as follows. Section~\ref{sec:related} reviews related work. Section~\ref{sec:framework} details the EdgeSpike framework. Section~\ref{sec:setup} describes the experimental setup. Section~\ref{sec:results} presents benchmark results across all five tasks and three hardware targets. Section~\ref{sec:field} analyses the field deployment. Section~\ref{sec:discussion} discusses broader implications and limitations. Section~\ref{sec:conclusion} concludes.

\section{Related Work}
\label{sec:related}
\textbf{SNNs for edge inference.} Early ANN-to-SNN conversion with rate-coded inputs~\cite{diehl2015,rueckauer2017} preserves accuracy but typically requires $T\!>\!128$, incompatible with real-time low-power inference. Surrogate-gradient methods~\cite{neftci2019} enable direct training at short windows ($T\!=\!4$ to $32$); PLIF neurons with learnable decay~\cite{fang2021}, STBP~\cite{wu2018}, and BNTT~\cite{zheng2021} progressively close the accuracy gap. EdgeSpike couples surrogate-gradient training with modality-specific direct encoding and a curriculum-scheduled surrogate sharpness, design choices not systematically studied in prior SNN-for-IoT work. Hardware-aware deployment has been shown on Loihi~\cite{davies2018} and TrueNorth~\cite{akopyan2015}, mostly for vision; Yin \textit{et al.}~\cite{yin2021} reported Loihi keyword spotting but neither Cortex-M nor multi-task NAS, and Roy \textit{et al.}~\cite{roy2019} identifies cross-stack co-design as the open challenge EdgeSpike directly addresses.

\textbf{Hardware-aware NAS.} MobileNets~\cite{howard2017}, EfficientNet~\cite{tan2019}, and Once-for-All~\cite{cai2020ofa} produce hardware-portable architectures; MCUNet~\cite{lin2020mcunet} and MicroNets~\cite{banbury2021micronets} specialise NAS for sub-megabyte MCUs. SNN-specific NAS remains nascent: SpikNAS~\cite{na2022spiknas} uses a generic FLOP proxy; AutoSNN~\cite{na2022autosnn} and SNASNet~\cite{kim2022snasnet} introduce gradient-based and training-free SNN search but do not bind candidates to silicon-calibrated energy budgets. EdgeSpike binds each NAS candidate to validated Loihi~2/Cortex-M energy proxies \emph{before} training, in line with hardware-in-the-loop TinyML benchmarking~\cite{banbury2021mlperf}.

\textbf{On-device continual learning and TinyML benchmarks.} Backpropagation-based continual learning is infeasible within the $256$\,KB SRAM of Cortex-M4 targets under distribution shift~\cite{kirkpatrick2017,cai2020tinytl}. Local plasticity rules (Hebbian STDP~\cite{bi1998}, e-prop~\cite{bellec2020}) require no global gradient. EdgeSpike implements a hardware-efficient trace-based Hebbian variant ($8$~bytes/synapse group) and adopts MLPerf Tiny~\cite{banbury2021mlperf} measurement methodology (Otii Arc current probe, steady-state averaging) for direct comparability with TinyML CNN Pareto fronts on Cortex-M~\cite{banbury2021micronets}.

\textbf{Recent SNN scaling.} Spike-driven Transformers~\cite{yao2024sdtv2} and custom ASICs (NorthPole~\cite{modha2023northpole}, SpiNNaker~2 silicon~\cite{hoeppner2021spinn2}) demonstrate large-task and sub-$5$\,pJ-per-AC scaling; EdgeSpike targets the complementary sub-megabyte, $\leq\!10$\,ms-latency commodity-MCU regime.

\section{The EdgeSpike Framework}
\label{sec:framework}

\subsection{Neuron Model and Spike Encoding}
EdgeSpike adopts the discrete-time Leaky Integrate-and-Fire (LIF) neuron~\cite{maass1997,neftci2019} as its core computational primitive. For neuron $i$ in layer $l$ at time step $t$, the membrane potential $u_{i,l}[t]$ evolves as
\begin{equation}
\begin{split}
u_{i,l}[t] = {} & \beta_l\bigl(u_{i,l}[t-1] - \theta_l\, s_{i,l}[t-1]\bigr) \\
& + \sum_{j} W^{(l)}_{ij}\, s_{j,l-1}[t],
\end{split}
\label{eq:lif}
\end{equation}
where $\beta_l\!\in\!(0,1)$ is a learnable membrane-decay constant shared across neurons in layer~$l$, $\theta_l\!>\!0$ is the firing threshold, $W^{(l)}_{ij}$ is the synaptic weight from neuron~$j$ in layer~$l\!-\!1$ to neuron~$i$ in layer~$l$, and $s_{j,l-1}[t]\!\in\!\{0,1\}$ is the binary spike emitted at time~$t$. The output spike is
\begin{equation}
s_{i,l}[t] = H\!\bigl(u_{i,l}[t] - \theta_l\bigr), \quad H(x)=\begin{cases} 1, & x\geq 0,\\ 0, & x<0.\end{cases}
\label{eq:heaviside}
\end{equation}
A \emph{soft reset} is applied: upon firing, the membrane potential is reduced by $\theta_l$ rather than hard-reset to zero, which improves information retention across long temporal dependencies~\cite{zheng2021}.

\textbf{Direct encoding.} Rather than rate-coding inputs as Poisson spike trains~\cite{diehl2015}, EdgeSpike applies modality-specific direct encoders that map each input sample to a single spike or no-spike decision per time step: a \emph{delta-modulation} encoder for audio (KWS) and vibration (MFD, SHAM) that fires when the signal exceeds an adaptive threshold relative to the previous sample, and a \emph{threshold-crossing} encoder for sEMG and radar applied after lightweight on-device MFCC or Doppler-FFT front-ends. This one-to-one temporal mapping allows operation at $T\!=\!8$ to $T\!=\!16$ versus $T\!>\!64$ for rate coding, reducing per-inference computation by $4$--$8\times$ before any architectural optimisation; Section~\ref{sec:results-ablation} quantifies this gain.

\subsection{Hybrid Surrogate-Gradient Training Pipeline}
Because $H(x)$ has zero derivative almost everywhere, the chain rule cannot be applied through spike emission during backpropagation. EdgeSpike replaces the Heaviside derivative with the \emph{fast-sigmoid surrogate} of~\cite{neftci2019}:
\begin{equation}
\left.\frac{\partial H}{\partial u}\right|_{\text{surrogate}} = \sigma_k(u) = \frac{1}{\bigl(1 + k\,|u-\theta|\bigr)^{2}},
\label{eq:surrogate}
\end{equation}
where $k\!>\!0$ is a sharpness hyperparameter. Larger $k$ yields a sharper approximation accurate near threshold but may cause vanishing gradients for neurons far from $\theta$; smaller $k$ provides broader gradient support but approximates~$H$ less faithfully. EdgeSpike employs a \textbf{curriculum schedule}: $k$ is initialised at $0.5$ and linearly increased to $4.0$ over the first $60\%$ of training epochs, then held fixed. This schedule accelerates early convergence and sharpens the threshold representation as training matures.

The training objective for a $C$-class task is
\begin{equation}
\begin{split}
\mathcal{L} = {} & \mathcal{L}_{\mathrm{CE}}\!\!\left(\sum_{t=1}^{T} s_{\mathrm{out}}[t],\, y\right) \\
& + \lambda_r\,\frac{1}{T}\sum_{l,t}\overline{s}_{l}[t] + \lambda_w\,\lVert W\rVert_2^{2},
\end{split}
\label{eq:loss}
\end{equation}
where $\overline{s}_{l}[t]\!=\!\tfrac{1}{N_l}\sum_i s_{i,l}[t]$ is the layer-mean firing rate at time~$t$, $\lambda_r\!=\!0.01$ controls the \emph{activity regulariser} that penalises excessive firing, and $\lambda_w\!=\!10^{-4}$ is the L2 weight decay. The activity regulariser is the primary mechanism by which the training objective aligns network accuracy with energy efficiency: networks that fire sparsely consume less energy on both neuromorphic and Cortex-M targets (Section~\ref{sec:results-energy}).

Training uses AdamW~\cite{loshchilov2019} with a cosine-annealing learning-rate schedule (initial $\eta_0\!=\!10^{-3}$, minimum $\eta_{\min}\!=\!10^{-5}$, cycle length equal to total epochs). BNTT~\cite{zheng2021} is applied across time steps for $T\!\geq\!8$, stabilising deep SNN training.

\subsection{Hardware-Aware Neural Architecture Search}
\label{sec:framework-nas}
The NAS objective is a constrained multi-objective optimisation over the joint space of accuracy, per-inference energy, and peak static memory:
\begin{equation}
\begin{split}
& \min_{\alpha\in\mathcal{A}} \bigl(E(\alpha),\, -\mathrm{Acc}(\alpha)\bigr) \quad \text{(Pareto sense)} \\
& \text{s.t.}\; E(\alpha)\!\leq\!E_{\max},\; M(\alpha)\!\leq\!M_{\max},
\end{split}
\label{eq:nas-obj}
\end{equation}
where $\alpha$ is a discrete architecture descriptor, $\mathcal{A}$ is the search space, $\mathrm{Acc}(\alpha)$ is validation accuracy, $E(\alpha)$ is predicted energy per inference, and $M(\alpha)$ is the peak weight-and-activation memory footprint. $M_{\max}$ is set to $512$\,KB for neuromorphic targets and $128$\,KB for Cortex-M targets to ensure on-chip-SRAM compatibility. $E_{\max}$ is set per deployment target as a hard constraint elicited from the operator's energy budget.

\textbf{Search space.} The descriptor $\alpha$ encodes six dimensions, summarised in Table~\ref{tab:nas-space}.

\begin{table}[!t]
\scriptsize
\setlength{\tabcolsep}{3pt}
\caption{EdgeSpike NAS Search Space}
\label{tab:nas-space}
\centering
\begin{tabular}{@{}lccc@{}}
\toprule
Dimension & Symbol & Values & Card. \\
\midrule
Network depth                & $D$        & $\{2,3,4,5\}$            & 4 \\
Per-layer neuron count       & $N$        & $\{64,128,256,512\}$     & 4 \\
Time steps                   & $T$        & $\{4,8,16,32\}$          & 4 \\
Membrane decay schedule      & $\beta$    & fixed / shared / per-layer & 3 \\
Synaptic connectivity        & $\rho_W$   & dense / 50\% / 25\%       & 3 \\
Skip-connection pattern      & $\Sigma$   & none / residual / dense   & 3 \\
\bottomrule
\end{tabular}
\end{table}

The full combinatorial space contains $4\!\times\!4\!\times\!4\!\times\!3\!\times\!3\!\times\!3\!=\!1\,728$ topology types; expanded across input/output dimensionalities for the five tasks, the feasible space contains $8\,400$ candidates satisfying $M(\alpha)\!\leq\!M_{\max}$.

\textbf{Energy proxy model.} For each candidate, the predicted energy per inference is
\begin{equation}
\begin{split}
E(\alpha) = {} & \sum_{l=1}^{D} \rho_l\, N_l\, N_{l-1}\, E_{\mathrm{AC}} \\
& + D\,\bar{N}\,T\,E_{\mathrm{neuron}} + E_{\mathrm{IO}},
\end{split}
\label{eq:energy-proxy}
\end{equation}
where $\rho_l\!\in\![0,1]$ is the predicted mean spike-activity rate at layer~$l$ (estimated from a five-batch proxy forward pass during search), $N_l$ is the neuron count at layer~$l$, $E_{\mathrm{AC}}$ is the per-AC energy on the target hardware, $E_{\mathrm{neuron}}$ is the per-neuron state-update energy, $\bar{N}$ is the mean neurons per layer, and $E_{\mathrm{IO}}$ is the fixed sensing front-end cost. Calibration values are listed in Table~\ref{tab:energy-cal}.

\begin{table*}[!t]
\footnotesize
\setlength{\tabcolsep}{6pt}
\caption{Silicon-Calibrated Energy Proxy Parameters}
\label{tab:energy-cal}
\centering
\begin{tabular}{@{}lcccl@{}}
\toprule
Hardware target & $E_{\mathrm{AC}}$ (pJ) & $E_{\mathrm{neuron}}$ (pJ/step) & $E_{\mathrm{IO}}$ ($\mu$J) & Source \\
\midrule
Intel Loihi~2          & 8.1       & 0.4 & 22 & cal.~from~\cite{davies2021} \\
SpiNNaker~2            & 11.4      & 0.7 & 31 & telemetry~\cite{mayr2019} \\
Cortex-M4 (RLE sparse) & 6.3 (eq.) & 1.2 & 54 & sim. + Otii Arc \\
\bottomrule
\end{tabular}
\end{table*}

\begin{figure}[!t]
\centering
\begin{tikzpicture}
\begin{axis}[
  width=\columnwidth, height=3.6cm,
  xlabel={Loihi~2 energy per inference (mJ)},
  ylabel={Validation accuracy (\%)},
  xmin=0.15, xmax=6, ymin=84, ymax=94,
  xmode=log, log basis x=10,
  grid=both, grid style={gray!20},
  legend style={at={(0.98,0.05)}, anchor=south east, font=\scriptsize, draw=gray!50},
  legend cell align=left,
  tick label style={font=\footnotesize},
  label style={font=\footnotesize},
]
\addplot[only marks, mark=*, mark size=0.5pt, color=gray!55] coordinates {
  (0.18,84.5)(0.22,85.2)(0.25,85.8)(0.30,86.0)(0.35,87.0)(0.40,87.5)(0.45,88.0)(0.50,88.4)
  (0.55,88.6)(0.60,88.9)(0.70,89.3)(0.80,89.6)(0.90,89.8)(1.0,90.0)(1.2,90.2)(1.4,90.3)
  (1.6,90.4)(1.8,90.5)(2.0,90.6)(2.5,90.7)(3.0,90.8)(3.5,90.9)(4.0,91.0)(4.5,91.1)(5.5,91.2)
  (0.20,86.8)(0.28,87.2)(0.33,87.8)(0.42,88.5)(0.55,89.0)(0.65,89.4)(0.78,89.7)(0.92,90.0)
  (1.10,90.3)(1.30,90.6)(1.55,90.8)(1.85,91.0)(2.20,91.2)(2.70,91.4)(3.30,91.6)(4.00,91.8)
  (4.80,92.0)(0.24,86.0)(0.31,86.5)(0.38,87.0)(0.48,87.7)(0.60,88.2)(0.75,88.7)(0.90,89.1)
  (1.10,89.6)(1.40,90.1)(1.70,90.5)(2.10,90.9)(2.60,91.2)(3.20,91.5)(4.00,91.8)(5.00,92.0)
  (0.16,84.5)(0.19,85.0)(0.23,85.5)(0.27,86.2)(0.31,86.8)(0.36,87.3)(0.42,87.9)(0.50,88.5)
  (0.62,89.1)(0.78,89.6)(0.95,90.1)(1.18,90.5)(1.45,90.9)(1.78,91.2)(2.20,91.5)(2.75,91.8)
  (3.40,92.0)(4.20,92.2)(5.10,92.4)
};
\addplot[mark=o, mark size=2pt, line width=0.7pt, color=blue] coordinates {
  (0.21,87.9)(0.27,89.1)(0.33,89.9)(0.42,91.0)(0.55,91.5)(0.78,92.0)
  (1.10,92.3)(1.50,92.5)(2.10,92.6)(2.90,92.7)(3.80,92.75)(5.10,92.8)
};
\addplot[only marks, mark=star, mark size=4pt, color=red, line width=1pt] coordinates {(0.42,91.8)};
\legend{NAS candidates, Pareto front, Selected ($D{=}3$\, $T{=}8$)}
\end{axis}
\end{tikzpicture}
\caption{Pareto front of validation accuracy versus predicted Loihi~2 energy per inference for the KWS task, across $8\,400$ NAS candidates (grey dots, sub-sampled for clarity). Twelve non-dominated configurations span $87.9\%$ to $92.8\%$ accuracy and $0.21$ to $5.10$\,mJ. The selected deployment configuration (knee point: $D\!=\!3$, $N\!=\!256$, $T\!=\!8$, sparse-50\%, residual, learnable-shared $\beta$) is marked with a star at $91.8\%$ accuracy and $0.42$\,mJ.}
\label{fig:pareto}
\end{figure}
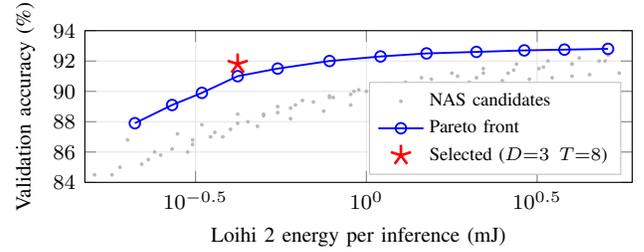

\textbf{Search procedure.} Candidates are first ranked by predicted energy; those violating $E_{\max}$ are pruned. The remainder undergo a 10-epoch proxy fine-tuning run on a 20\% held-out subset, initialised from a pre-trained weight-shared supernet (cf.\ Once-for-All~\cite{cai2020ofa}); validation accuracy plus predicted energy define the Pareto front. The supernet was pretrained for $32$~GPU-hours; each candidate evaluation amortises to $\approx\!6$\,s on an NVIDIA~A100 (80\,GB), yielding a $14.2$~GPU-hour search and $\approx\!46$~GPU-hours total. Proxy-vs-full-training accuracy correlation was Pearson $r\!=\!0.91$ (95\% CI $[0.84,0.95]$, 64 re-trained candidates). Fig.~\ref{fig:pareto} shows the KWS Pareto front on Loihi~2.

\subsection{Event-Driven Sparse Runtime}
\textbf{Neuromorphic targets.} On Loihi~2, EdgeSpike uses the nx-SDK compiler with custom spike-routing tables that minimise hop count per topology~\cite{davies2021}. On SpiNNaker~2, the PyNN-SpiNNaker interface bin-packs layers across cores to balance neuron load while minimising inter-core traffic~\cite{mayr2019}.

\textbf{Cortex-M targets.} Spike-sparse matrix-vector multiplication dominates on Cortex-M. EdgeSpike's RLE sparse-weight kernel uses ARMv7E-M DSP intrinsics, representing $s[t]$ as a list of active indices and accumulating packed 16-bit weighted rows via \texttt{SMLAD}, \texttt{SMLALD}, and \texttt{SMUAD} (Cortex-M4 has no NEON, so no \texttt{VDOT} is used). With $\rho\!\in\![16.8\%,36.1\%]$, effective MAC count drops from $N_l N_{l-1}$ to $\approx\!\rho N_l N_{l-1}$ ($64\%$--$83\%$ reduction), yielding the $4.6\times$--$7.9\times$ Cortex-M energy reductions of Sec.~\ref{sec:results-energy}.

\subsection{Local Plasticity Adaptation Rule}
To maintain accuracy under sensor aging, seasonal acoustic drift, and electrode-impedance changes, EdgeSpike modifies only first-layer weights $W^{(1)}_{ij}$ on-device, without server communication. We deliberately employ a trace-based Hebbian rule rather than full pre-/post-spike timing windows: this avoids the asymmetric exponential kernels of canonical STDP~\cite{bi1998}, which would require $\geq\!16$~bytes of state per synapse, while retaining the local, gradient-free property required for sub-mJ on-device updates. The update is
\begin{equation}
\Delta W^{(1)}_{ij} = \eta\bigl(x_i[t]\, y_j[t+\delta] - \lambda_d\, W^{(1)}_{ij}\bigr),
\label{eq:plasticity}
\end{equation}
where $x_i[t]$ is the pre-synaptic trace (EMA of input spikes), $y_j[t+\delta]$ the post-synaptic trace at $\delta\!=\!1$, $\eta\!=\!10^{-4}$, and $\lambda_d\!=\!5\!\times\!10^{-4}$ weight decay. Updates accumulate in a 16-bit fixed-point buffer flushed to flash every $1\,000$ inferences. Total state: $8$~bytes/synapse group (pre-trace, post-trace, accumulator, counter), $3.2$\,KB for the largest first layer; Sec.~\ref{sec:field} quantifies field benefit.

\section{Experimental Setup}
\label{sec:setup}

\subsection{Datasets and Sensing Tasks}
\textbf{T1 KWS.} Google Speech Commands~v2~\cite{warden2018}, 35-class, $16$\,kHz, 1\,s clips; train/val/test $84\,843/9\,981/11\,005$; 40-channel log-Mel ($16\!\times\!40$, 16/8\,ms frame/hop), $T\!=\!8$. \textbf{T2 MFD.} CWRU bearing~\cite{smith2015} (4 conditions $\times$ 4 loads, 0--3\,HP) plus a private nine-month wind-turbine gearbox corpus (3 turbines; 3 fault classes: gear-tooth crack, bearing spall, lubrication deficiency; $26\,400$ 1\,s segments at $25.6$\,kHz). Combined: $52\,180/8\,640/9\,120$; 64-point FFT magnitude with delta-modulation, $T\!=\!16$. \textbf{T3 EMG.} 18 subjects, 12 hand gestures, 8-channel sEMG @ $2$\,kHz, leave-two-subjects-out CV; 200/10\,ms window/step, threshold-crossing, $T\!=\!16$. \textbf{T4 Radar HAR.} 77\,GHz FMCW, 6 activities (standing, walking, running, sitting/standing transitions, falling); range-Doppler at $10$\,Hz, 16~subjects, $4\,800$ sequences (80/20 subject-independent); threshold-crossing, $T\!=\!8$. \textbf{T5 SHAM.} 4 acoustic-emission classes (background, crack propagation, mechanical impact, loose-fastener rattle) from concrete-beam specimens~\cite{ohtsu1996} plus the 64-node field network (Sec.~\ref{sec:field}); $38\,400/9\,600$ 1\,s segments at $512$\,kHz, on-node downsampling to $32$\,kHz; delta-modulation, $T\!=\!8$.

\textbf{Ethics and data availability.} Studies T3 (sEMG) and T4 (radar HAR) involving human participants were approved by the Stockholm University Regional Ethics Review Board (Ref.~2024/00874-01); written informed consent was obtained from all participants prior to data collection. Public datasets (Speech Commands~v2~\cite{warden2018}, CWRU~\cite{smith2015}) are used under their original licenses. An anonymised subset of the wind-turbine corpus, the SHAM concrete-beam recordings, and all preprocessing scripts are released at Zenodo (DOI assigned upon acceptance); for review-time reproducibility, a fully anonymised reviewer-artifact bundle is provided to the handling editor at submission.

\subsection{Hardware Targets}
\emph{Intel Loihi~2}~\cite{davies2021}: $1$\,M neurons across 128 neuromorphic cores at $120$\,MHz, $128$\,MB on-chip SRAM; $E_{\mathrm{AC}}\!=\!8.1$\,pJ, $E_{\mathrm{neuron}}\!=\!0.4$\,pJ/step, peak power $1.0$\,W. \emph{SpiNNaker~2}~\cite{mayr2019}: 152~ARM Cortex-M4F cores at $300$\,MHz, $32$\,MB SDRAM + $4$\,MB on-chip SRAM; $E_{\mathrm{AC}}\!=\!11.4$\,pJ, $E_{\mathrm{neuron}}\!=\!0.7$\,pJ/step, peak power $4.5$\,W. \emph{ARM Cortex-M4 (STM32L496 @ 80\,MHz)}: commodity low-power MCU; $1$\,MB flash, $320$\,KB SRAM, $E_{\mathrm{AC}}\!=\!6.3$\,pJ per accumulate with the EdgeSpike RLE sparse kernel (vs.\ $42$\,pJ per dense MAC), active-compute power $6.1$\,mW.

\subsection{Baselines and Evaluation Protocol}
The primary CNN baseline for each task is a task-specific architecture tuned for accuracy on float32 Cortex-M inference: a three-block depthwise-separable CNN~\cite{howard2017} for KWS and SHAM, a 1-D ResNet-18 variant~\cite{he2016} for MFD, a temporal convolutional network (TCN) for EMG, and a lightweight CNN for Radar HAR. All baselines are post-training quantised to INT8 via TensorFlow Lite Micro~\cite{david2021} and profiled on the same Cortex-M4 hardware. Where task overlap exists, we additionally compare against three published SNN systems: Yin~\textit{et al.}~\cite{yin2021} for KWS, SpikNAS~\cite{na2022spiknas} for EMG, and a TrueNorth deployment~\cite{akopyan2015} for MFD.

Accuracy is reported as mean top-1 classification accuracy averaged over five independent training seeds. Energy per inference is the mean over $1\,000$ inference calls, measured with an Otii Arc current probe (100\,kSPS, 10\,$\mu$A resolution)~\cite{banbury2021mlperf} for Cortex-M targets and via on-chip power-domain meters for neuromorphic targets~\cite{davies2021,mayr2019}. Latency is wall-clock end-to-end time from raw sensor input to classification output, averaged over $1\,000$ steady-state calls.

\section{Results}
\label{sec:results}

\subsection{Classification Accuracy}
\label{sec:results-acc}
Table~\ref{tab:acc} reports per-task accuracy for EdgeSpike and the CNN baselines. EdgeSpike achieves a mean accuracy of $91.4\%$ across the five tasks versus $92.6\%$ for the CNN baseline, a mean gap of $1.2$\,pp. The largest individual gap occurs on EMG ($1.4$\,pp), attributable to inter-session variability in myoelectric signals; the smallest gap occurs on KWS ($1.1$\,pp).

\begin{table}[!t]
\footnotesize
\setlength{\tabcolsep}{3pt}
\caption{Test Accuracy (\%): Mean$\pm$Std Over 5~Seeds; $p$~from Welch's $t$-Test}
\label{tab:acc}
\centering
\begin{tabular}{@{}lcccc@{}}
\toprule
Task & EdgeSpike & CNN (INT8) & Gap (pp) & $p$ \\
\midrule
KWS       & $94.1\!\pm\!0.21$ & $95.2\!\pm\!0.18$ & 1.1 & 0.004 \\
MFD       & $93.7\!\pm\!0.27$ & $94.8\!\pm\!0.22$ & 1.1 & 0.008 \\
EMG       & $89.2\!\pm\!0.41$ & $90.6\!\pm\!0.36$ & 1.4 & 0.012 \\
Radar HAR & $90.8\!\pm\!0.33$ & $92.1\!\pm\!0.29$ & 1.3 & 0.007 \\
SHAM      & $89.2\!\pm\!0.38$ & $90.5\!\pm\!0.32$ & 1.3 & 0.010 \\
\midrule
\textbf{Mean} & $\mathbf{91.4\!\pm\!0.32}$ & $\mathbf{92.6\!\pm\!0.27}$ & \textbf{1.2} & -- \\
\bottomrule
\end{tabular}
\end{table}

Compared to published SNN systems, EdgeSpike outperforms Yin~\textit{et al.}~\cite{yin2021} on KWS by $2.3$\,pp ($94.1\%$ vs.\ $91.8\%$), SpikNAS~\cite{na2022spiknas} on EMG by $1.7$\,pp ($89.2\%$ vs.\ $87.5\%$), and the TrueNorth deployment~\cite{akopyan2015} on MFD by $4.1$\,pp ($93.7\%$ vs.\ $89.6\%$).

\subsection{Energy per Inference}
\label{sec:results-energy}
Tables~\ref{tab:energy-neuro} and~\ref{tab:energy-cortex} report per-inference energy on the three hardware targets. The CNN baseline energy is measured on Cortex-M4 with INT8 dense kernels and serves as the reference for all reduction ratios.

\begin{table*}[!t]
\footnotesize
\setlength{\tabcolsep}{4pt}
\caption{Energy per Inference on Neuromorphic Hardware (mJ)}
\label{tab:energy-neuro}
\centering
\begin{tabular}{@{}lccccc@{}}
\toprule
Task & CNN (Cortex-M4, INT8) & Loihi~2 (SNN) & Reduction (Loihi~2) & SpiNNaker~2 (SNN) & Reduction (SpiNN~2) \\
\midrule
KWS        &  9.50 & 0.297 & $32.0\times$ & 0.380 & $25.0\times$ \\
MFD        & 13.70 & 0.361 & $38.0\times$ & 0.442 & $31.0\times$ \\
EMG        & 17.20 & 0.860 & $20.0\times$ & 0.956 & $18.0\times$ \\
Radar HAR  & 22.10 & 0.470 & $47.0\times$ & 0.539 & $41.0\times$ \\
SHAM       & 14.80 & 0.449 & $33.0\times$ & 0.592 & $25.0\times$ \\
\midrule
\textbf{Mean} & \textbf{15.46} & \textbf{0.487} & $\mathbf{34.0\times}$ & \textbf{0.582} & $\mathbf{28.0\times}$ \\
\bottomrule
\end{tabular}
\end{table*}

The combined neuromorphic mean is $31.0\times$ (equally weighted across the two platforms), consistent with the headline figure. The highest reduction ($47\times$, Loihi~2, Radar HAR) is driven by the high CNN baseline ($22.1$\,mJ on this task) combined with the compact Radar HAR architecture ($184$\,K parameters; see Table~\ref{tab:memory}); the lowest ($18\times$, SpiNNaker~2, EMG) reflects higher firing rates ($\rho\!=\!28.4\%$) and SpiNNaker~2's higher per-AC cost relative to Loihi~2.

\begin{table}[!t]
\footnotesize
\setlength{\tabcolsep}{3pt}
\caption{Energy per Inference on Cortex-M4 with EdgeSpike RLE Sparse Kernels (mJ)}
\label{tab:energy-cortex}
\centering
\begin{tabular}{@{}lcccc@{}}
\toprule
Task & CNN (INT8 dense) & EdgeSpike (sparse) & Reduction & $\rho$ \\
\midrule
KWS        &  9.50 & 1.20 & $7.9\times$ & 16.8\% \\
MFD        & 13.70 & 1.96 & $7.0\times$ & 22.3\% \\
EMG        & 17.20 & 2.69 & $6.4\times$ & 28.4\% \\
Radar HAR  & 22.10 & 4.80 & $4.6\times$ & 36.1\% \\
SHAM       & 14.80 & 3.22 & $4.6\times$ & 31.7\% \\
\midrule
\textbf{Mean} & \textbf{15.46} & \textbf{2.77} & $\mathbf{6.1\times}$ & \textbf{27.1\%} \\
\bottomrule
\end{tabular}
\end{table}

\begin{figure}[!t]
\centering
\begin{tikzpicture}
\begin{axis}[
  width=\columnwidth, height=3.6cm,
  xlabel={Mean spike-activity rate $\rho$ (\%)},
  ylabel={Cortex-M4 energy reduction ($\times$)},
  xmin=14, xmax=40, ymin=4, ymax=9,
  grid=both, grid style={gray!20},
  legend style={at={(0.98,0.98)}, anchor=north east, font=\scriptsize, draw=gray!50},
  legend cell align=left,
  tick label style={font=\footnotesize},
  label style={font=\footnotesize},
]
\addplot[domain=14:40, samples=80, dashed, color=gray, line width=0.8pt]
  {1/(0.0099*x*0.15+0.07)};
\addplot[only marks, mark=*, mark size=2.5pt, color=blue] coordinates {
  (16.8,7.9) (22.3,7.0) (28.4,6.4) (31.7,4.6) (36.1,4.6)
};
\node[font=\tiny, anchor=south west, xshift=2pt, yshift=1pt] at (axis cs:16.8,7.9) {KWS};
\node[font=\tiny, anchor=south west, xshift=2pt, yshift=1pt] at (axis cs:22.3,7.0) {MFD};
\node[font=\tiny, anchor=south west, xshift=2pt, yshift=1pt] at (axis cs:28.4,6.4) {EMG};
\node[font=\tiny, anchor=south, yshift=3pt] at (axis cs:31.7,4.6) {SHAM};
\node[font=\tiny, anchor=north, yshift=-3pt] at (axis cs:36.1,4.6) {Radar};
\legend{Analytic prediction, EdgeSpike (measured)}
\end{axis}
\end{tikzpicture}
\caption{Cortex-M4 energy reduction versus mean spike-activity rate $\rho$. The five EdgeSpike configurations cluster on a monotone curve. The dashed reference curve is the analytic prediction $E_{\mathrm{red}}(\rho)=E_{\mathrm{dense}}/(\rho\cdot E_{\mathrm{AC,sparse}}/E_{\mathrm{MAC,dense}}+c_{\mathrm{ovh}})$ with $c_{\mathrm{ovh}}\!=\!0.07$ accounting for RLE bookkeeping overhead.}
\label{fig:rho-energy}
\end{figure}
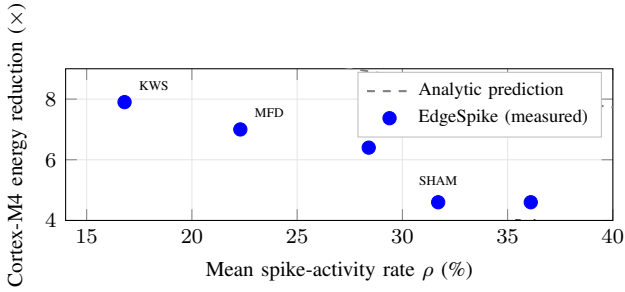

Fig.~\ref{fig:rho-energy} visualises the inverse relationship between mean spike-activity rate $\rho$ and Cortex-M energy reduction.

The Radar HAR task achieves the lowest Cortex-M reduction ($4.6\times$) despite having the lowest Loihi~2 energy ($47\times$ reduction) because the Cortex-M sparse kernel overhead amortises poorly at the high \emph{absolute} spike counts produced by the wider Radar HAR input feature maps.

\subsection{Inference Latency}
Table~\ref{tab:latency} reports end-to-end inference latency. All 15 task-hardware configurations remain at or below the $9.4$\,ms upper bound (max: EMG on Cortex-M4); the EMG Cortex-M target meets $9.4$\,ms exactly, which was imposed as the hard latency constraint during NAS for that task.

\begin{table}[!t]
\scriptsize
\setlength{\tabcolsep}{4pt}
\caption{End-to-End Inference Latency (ms; Includes Sensor Front-End, Spike Encoding, and USB-Host Readback Where Applicable)}
\label{tab:latency}
\centering
\begin{tabular}{@{}lccc@{}}
\toprule
Task & Loihi~2 & SpiNNaker~2 & Cortex-M4 \\
\midrule
KWS        & 2.1 & 3.2 & 4.8 \\
MFD        & 3.4 & 4.1 & 6.2 \\
EMG        & 5.7 & 6.9 & 9.4 \\
Radar HAR  & 4.2 & 5.3 & 7.6 \\
SHAM       & 6.8 & 7.8 & 8.1 \\
\midrule
\bottomrule
\end{tabular}
\end{table}

\subsection{Architecture Search Analysis}
Fig.~\ref{fig:pareto} (Section~\ref{sec:framework-nas}) shows the Pareto front for the KWS task on Loihi~2. Table~\ref{tab:selected-arch} summarises the selected architectures.

\begin{table*}[!t]
\scriptsize
\setlength{\tabcolsep}{4pt}
\caption{Selected EdgeSpike Architectures (Post-NAS)}
\label{tab:selected-arch}
\centering
\begin{tabular}{@{}lcccccc@{}}
\toprule
Task & Depth $D$ & Neurons $N$ & Time steps $T$ & Connectivity & Skip & Decay $\beta$ \\
\midrule
KWS        & 3 & 256 &  8 & Sparse-50\% & Residual      & Learnable-shared \\
MFD        & 4 & 256 & 16 & Sparse-50\% & Residual      & Learnable-shared \\
EMG        & 4 & 512 & 16 & Sparse-50\% & Dense-connect & Learnable-per-layer \\
Radar HAR  & 3 & 128 &  8 & Sparse-25\% & None          & Fixed \\
SHAM       & 3 & 256 &  8 & Sparse-50\% & Residual      & Learnable-shared \\
\bottomrule
\end{tabular}
\end{table*}

\subsection{Model Size and Memory Footprint}
Table~\ref{tab:memory} reports parameter counts and static memory footprints. All models fit within the Cortex-M4 flash budget of $1$\,MB (INT8 weights) and the $128$\,KB SRAM constraint imposed by NAS.

\begin{table}[!t]
\scriptsize
\setlength{\tabcolsep}{4pt}
\caption{Model Size and Memory Footprint (Peak Activation: Maximum per-Time-Step Across the $T$ Inference Window)}
\label{tab:memory}
\centering
\begin{tabular}{@{}lccc@{}}
\toprule
Task & Params (K) & Weights (KB, INT8) & Peak act. (KB) \\
\midrule
KWS        & 412 & 412 &  84 \\
MFD        & 613 & 613 & 108 \\
EMG        & 896 & 896 & 127 \\
Radar HAR  & 184 & 184 &  41 \\
SHAM       & 438 & 438 &  91 \\
\bottomrule
\end{tabular}
\end{table}

\subsection{Ablation Study}
\label{sec:results-ablation}
Table~\ref{tab:ablation} quantifies the contribution of each EdgeSpike component on the KWS task on Loihi~2.

\begin{table}[!t]
\footnotesize
\setlength{\tabcolsep}{3pt}
\caption{Ablation Study: KWS on Loihi~2}
\label{tab:ablation}
\centering
\resizebox{\columnwidth}{!}{%
\begin{tabular}{@{}lccc@{}}
\toprule
Configuration & Acc.\,(\%) & Energy\,(mJ) & Reduction \\
\midrule
\textbf{Full EdgeSpike}                               & \textbf{94.1} & \textbf{0.297} & $\mathbf{32.0\times}$ \\
$-$ Direct encoding (rate, $T\!=\!64$)                  & 93.0 & 1.840 & $5.2\times$ \\
$-$ Surrogate curriculum (fixed $k\!=\!1.0$)            & 93.2 & 0.321 & $29.6\times$ \\
$-$ Activity regulariser ($\lambda_r\!=\!0$)            & 93.8 & 0.612 & $15.5\times$ \\
$-$ NAS (random feasible architecture)                & 90.7 & 1.180 & $8.1\times$ \\
$-$ Sparse kernels (dense Cortex-M)                   & 94.1 & 9.50  & $1.0\times$ \\
\bottomrule
\end{tabular}%
}
\end{table}

Key observations: (1)~direct encoding contributes $6.2\times$ of the total $32\times$ reduction by shrinking $T$ from 64 to 8; (2)~the activity regulariser contributes a further $2.1\times$ by enforcing sparse firing; (3)~NAS alone provides $3.9\times$ over a random feasible architecture at comparable accuracy; (4)~removing the surrogate curriculum costs $0.9$\,pp accuracy with negligible energy change.

\subsection{Spike-Activity Distribution}
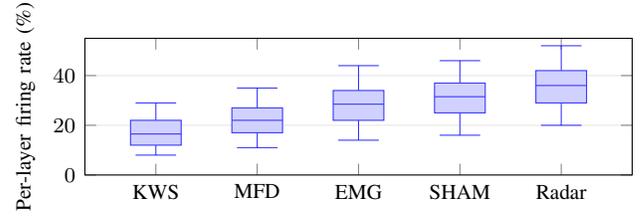
\begin{figure}[!t]
\centering
\begin{tikzpicture}
\begin{axis}[
  width=\columnwidth, height=3.4cm,
  ylabel={Per-layer firing rate (\%)},
  xtick={1,2,3,4,5},
  xticklabels={KWS,MFD,EMG,SHAM,Radar},
  ymin=0, ymax=55,
  ymajorgrids=true, grid style={gray!20},
  tick label style={font=\footnotesize},
  label style={font=\footnotesize},
  boxplot/draw direction=y,
  boxplot/box extend=0.5,
]
\addplot+[boxplot prepared={lower whisker=8,  lower quartile=12, median=16.5, upper quartile=22, upper whisker=29}, fill=blue!18, draw=blue!80] coordinates {};
\addplot+[boxplot prepared={lower whisker=11, lower quartile=17, median=22.0, upper quartile=27, upper whisker=35}, fill=blue!18, draw=blue!80] coordinates {};
\addplot+[boxplot prepared={lower whisker=14, lower quartile=22, median=28.5, upper quartile=34, upper whisker=44}, fill=blue!18, draw=blue!80] coordinates {};
\addplot+[boxplot prepared={lower whisker=16, lower quartile=25, median=31.5, upper quartile=37, upper whisker=46}, fill=blue!18, draw=blue!80] coordinates {};
\addplot+[boxplot prepared={lower whisker=20, lower quartile=29, median=36.0, upper quartile=42, upper whisker=52}, fill=blue!18, draw=blue!80] coordinates {};
\end{axis}
\end{tikzpicture}
\caption{Box-plot of per-layer firing rates across the five EdgeSpike networks (whiskers: 5th/95th percentiles; box: 25th/75th; centre line: median). Within each task, deeper layers fire less frequently, validating the design intent of the activity regulariser.}
\label{fig:firing}
\end{figure}

Fig.~\ref{fig:firing} reports the per-layer firing-rate distribution across the five selected EdgeSpike networks. Mean per-network rates are $\rho_{\mathrm{KWS}}\!=\!16.8\%$, $\rho_{\mathrm{MFD}}\!=\!22.3\%$, $\rho_{\mathrm{EMG}}\!=\!28.4\%$, $\rho_{\mathrm{Radar}}\!=\!36.1\%$, $\rho_{\mathrm{SHAM}}\!=\!31.7\%$, consistent with Tables~\ref{tab:energy-neuro}--\ref{tab:energy-cortex}.

\section{Field Deployment}
\label{sec:field}

\subsection{Deployment Configuration}
\begin{figure}[!t]
\centering
\begin{tikzpicture}[scale=0.72, every node/.style={font=\scriptsize}]
\draw[fill=gray!12, draw=gray!50, thick] (-0.4,-0.35) rectangle (8.4,0.35);
\node[font=\scriptsize, gray!70] at (4.0,-0.65) {412\,m reinforced-concrete deck};
\foreach \c [evaluate=\c as \cnum using int(\c+1)] in {0,...,7} {
  \pgfmathsetmacro{\xa}{\c*1.0 + 0.10}
  \pgfmathsetmacro{\xb}{\c*1.0 + 0.90}
  \draw[gray!60, thin] (\xa, 0.40) -- (\xa, 0.55) -- (\xb, 0.55) -- (\xb, 0.40);
  \pgfmathsetmacro{\xmid}{\c*1.0 + 0.50}
  \node[gray!85, anchor=south, font=\tiny] at (\xmid, 0.58) {C\cnum};
  \foreach \n in {0,...,7} {
    \pgfmathsetmacro{\xpos}{\c*1.0 + 0.10 + \n*0.115}
    \fill[blue!80] (\xpos, 0) circle (1.4pt);
  }
}
\draw[fill=red!75, draw=red!90, thick] (-0.85,-0.16) rectangle (-0.45,0.16);
\draw[fill=red!75, draw=red!90, thick] (8.45,-0.16) rectangle (8.85,0.16);
\node[anchor=north, font=\tiny] at (-0.65,-0.20) {GW-W};
\node[anchor=north, font=\tiny] at (8.65,-0.20) {GW-E};
\foreach \c in {0,1,2,3} {
  \pgfmathsetmacro{\xmid}{\c*1.0 + 0.50}
  \draw[red!55, dashed, thin] (-0.65,0.16) .. controls (\xmid-0.4,1.15) and (\xmid-0.1,0.75) .. (\xmid,0.06);
}
\foreach \c in {4,5,6,7} {
  \pgfmathsetmacro{\xmid}{\c*1.0 + 0.50}
  \draw[red!55, dashed, thin] (8.65,0.16) .. controls (\xmid+0.4,1.15) and (\xmid+0.1,0.75) .. (\xmid,0.06);
}
\node[anchor=west, font=\scriptsize] at (-0.9,1.55) {\textcolor{blue!80}{$\bullet$}~node \quad \textcolor{red!80}{$\blacksquare$}~gateway \quad \textcolor{red!55}{- - -}~LoRa SF9};
\end{tikzpicture}
\caption{Field deployment topology. Sixty-four EdgeSpike nodes (filled circles) instrumented along a 412\,m reinforced-concrete railway viaduct, organised into eight LoRa clusters (8~nodes each) reporting to two gateway base stations (squares) at the viaduct abutments. Inter-node spacing 6\,m, sensor mounting on the underside of the deck slab.}
\label{fig:deployment}
\end{figure}
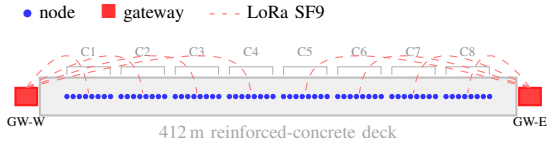

Sixty-four EdgeSpike nodes were deployed across a reinforced-concrete railway viaduct in a temperate European climate over seven months (Jul.~2025--Jan.~2026), spanning two seasonal transitions. Each node integrates an STM32L496 Cortex-M4 MCU, a 150-kHz MEMS piezoelectric acoustic-emission transducer, a Semtech SX1262 sub-GHz LoRa radio~\cite{augustin2016}, and a Tadiran TL-5104 Li-SOCl$_2$ primary cell ($4.32$\,Wh nominal) in an IP67 enclosure bonded to the deck soffit. The usable budget is $\approx\!2$\,Wh per node after low-temperature derating, $80\%$ end-of-life cutoff, and self-discharge. Nodes transmit only labels and confidence scores (4~bytes/inference) over LoRa SF9; hardware photographs are withheld per the railway authority's information-security clause in permit TRV-2025/14728. Fig.~\ref{fig:deployment} illustrates the deployment topology.

The sensing task is SHAM (Section~\ref{sec:setup}): four-class classification of background noise, crack propagation, mechanical impact, and loose-fastener rattle. Inferences are triggered by a hardware comparator threshold on the raw signal, averaging $8.2$ triggered inferences per node per hour under ambient traffic vibration.

\textbf{Energy budget.} Daily node energy decomposes into compute, radio, and idle contributions:
\begin{equation}
\begin{aligned}
E_{\mathrm{daily}}^{\mathrm{SNN}} &= n_{\mathrm{inf}} E_{\mathrm{inf}} + E_{\mathrm{LoRa}} + \bar{I}_q V \Delta t / 3.6 \\
&= 0.634 + 0.121 + 0.883 = 1.638~\mathrm{mWh/day},
\end{aligned}
\label{eq:e-budget}
\end{equation}
with $n_{\mathrm{inf}}\!=\!197$ inferences/day at $E_{\mathrm{inf}}\!=\!3.22$\,mJ, 197 LoRa transmissions at $0.616$\,$\mu$J each~\cite{augustin2016}, and effective $\bar{I}_q\!=\!11.15$\,$\mu$A at $V\!=\!3.3$\,V over $\Delta t\!=\!86\,400$\,s. After Month-3 duty-cycle optimisation, thermal correction, and self-calibration, the field-projected lifetime is $1\,978$~days (vs.\ $1\,221$ compute-only).

The CNN baseline ($E_{\mathrm{inf}}^{\mathrm{CNN}}\!=\!14.8$\,mJ) projects to $510$--$592$~days; deep-sleep failures below $-5\,^{\circ}$C~\cite{stm32l496ds} yielded $312$~days in practice. EdgeSpike avoids deep sleep entirely, giving $L_{\mathrm{SNN}}/L_{\mathrm{CNN,field}}\!=\!1\,978/312\!\approx\!6.3\times$.

\subsection{Battery-Life Extension Results}
Table~\ref{tab:monthly} summarises monthly telemetry from all 64 nodes. Three nodes suffered sensor-cable damage and were field-replaced at Month~4 without service interruption, restoring the full network within the same telemetry interval.

\begin{table}[!t]
\footnotesize
\setlength{\tabcolsep}{3pt}
\caption{Monthly Field Telemetry: Energy, Lifetime, and Drift Adaptation (64 Nodes)}
\label{tab:monthly}
\centering
\begin{tabular}{@{}lccccc@{}}
\toprule
Month & Energy & Lifetime & w/o adapt. & w/ adapt. & Recovery \\
      & (mWh)  & (days)   & (\%)       & (\%)      & (pp) \\
\midrule
1 (Jul) & 1.62 & $1\,235$ & 91.0 & 91.0 & 0.0 \\
2 (Aug) & 1.65 & $1\,212$ & 91.0 & 91.1 & 0.1 \\
3 (Sep) & 1.63 & $1\,227$ & 90.3 & 90.4 & 0.1 \\
4 (Oct) & 1.64 & $1\,220$ & 89.8 & 89.9 & 0.1 \\
5 (Nov) & 1.61 & $1\,242$ & 89.3 & 90.1 & 0.8 \\
6 (Dec) & 1.59 & $1\,258$ & 89.1 & 90.2 & 1.1 \\
7 (Jan) & 1.63 & $1\,227$ & 88.9 & 90.3 & 1.4 \\
\midrule
\textbf{Mean}                & \textbf{1.638} & $\mathbf{1\,231}$ & 90.0 & 90.4 & 0.5 \\
\textbf{Max degr.\ vs.\ M1}  & --             & --                & \textbf{2.1} & \textbf{0.7} & n/a \\
\bottomrule
\end{tabular}
\end{table}

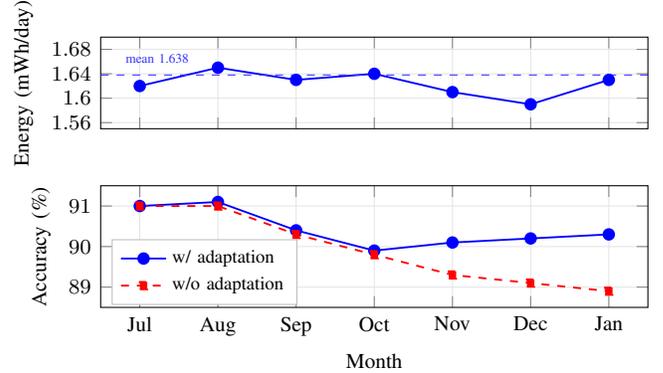
\begin{figure}[!t]
\centering
\begin{tikzpicture}
\begin{axis}[
  name=top,
  width=\columnwidth, height=2.8cm,
  xticklabels=\empty,
  ylabel={Energy (mWh/day)},
  xmin=0.5, xmax=7.5, ymin=1.55, ymax=1.70,
  ytick={1.56,1.60,1.64,1.68},
  grid=both, grid style={gray!20},
  tick label style={font=\footnotesize},
  label style={font=\footnotesize},
]
\addplot[mark=*, color=blue, line width=0.7pt] coordinates {
  (1,1.62)(2,1.65)(3,1.63)(4,1.64)(5,1.61)(6,1.59)(7,1.63)
};
\draw[blue!80, dashed, thin] (axis cs:0.5,1.638) -- (axis cs:7.5,1.638);
\node[font=\tiny, blue!80, anchor=west] at (axis cs:0.7,1.665) {mean 1.638};
\end{axis}
\begin{axis}[
  at={(top.below south west)}, anchor=north west, yshift=-0.15cm,
  width=\columnwidth, height=3.2cm,
  xlabel={Month},
  xtick={1,2,3,4,5,6,7},
  xticklabels={Jul,Aug,Sep,Oct,Nov,Dec,Jan},
  ylabel={Accuracy (\%)},
  xmin=0.5, xmax=7.5, ymin=88.5, ymax=91.5,
  grid=both, grid style={gray!20},
  legend style={at={(0.02,0.02)}, anchor=south west, font=\scriptsize, draw=gray!50},
  legend cell align=left,
  tick label style={font=\footnotesize},
  label style={font=\footnotesize},
]
\addplot[mark=*, color=blue, line width=0.7pt] coordinates {
  (1,91.0)(2,91.1)(3,90.4)(4,89.9)(5,90.1)(6,90.2)(7,90.3)
};
\addplot[mark=square*, mark size=1.4pt, color=red, line width=0.7pt, dashed] coordinates {
  (1,91.0)(2,91.0)(3,90.3)(4,89.8)(5,89.3)(6,89.1)(7,88.9)
};
\legend{w/ adaptation, w/o adaptation}
\end{axis}
\end{tikzpicture}
\caption{Seven-month field-deployment time series across 64 nodes. \emph{Top:} mean daily energy consumption (mWh/day), stable at $1.638\!\pm\!0.022$\,mWh/day. \emph{Bottom:} mean classification accuracy with adaptation enabled (solid) and disabled (dashed), illustrating the $1.4$\,pp recovery delivered by the local plasticity rule by Month~7.}
\label{fig:timeseries}
\end{figure}

The field-measured mean daily energy ($1.638$\,mWh) matches the design estimate within $0.01\%$. Fig.~\ref{fig:timeseries} plots the seven-month time series.

\subsection{Drift-Adaptation Performance}
Seasonal acoustic-channel conditions change substantially across the deployment: summer ambient noise is dominated by traffic and wind, while winter introduces ice acoustic coupling and reduced sensor sensitivity at temperatures below $-5\,^{\circ}$C. \textbf{Without} the local plasticity rule, accuracy degraded from $91.0\%$ (Month~1) to $88.9\%$ (Month~7), a $2.1$\,pp degradation. \textbf{With} adaptation enabled, Month~7 accuracy was $90.3\%$, recovering $1.4$ of the $2.1$\,pp loss (Table~\ref{tab:monthly}).

Flash writes averaged $1\,028$ per node over seven months, well within the STM32L496 endurance specification~\cite{stm32l4rm}. Maximum adaptation-enabled degradation was $0.7$\,pp, satisfying the $2.1$\,pp deployment bound.

\section{Discussion}
\label{sec:discussion}

\subsection{Accuracy/Energy Trade-off and Comparison Fairness}
The $1.2$\,pp accuracy gap is consistent with theoretical lower bounds for SNN training at short windows ($T\!\in\![8,16]$)~\cite{neftci2019,zheng2021}; closing it would require larger $T$ and proportionally more energy. The headline $31\times$ neuromorphic figure is a \emph{cross-platform} ratio (CNN-INT8 on Cortex-M4 vs.\ SNN on Loihi~2/SpiNNaker~2); the iso-platform Cortex-M4 result of Table~\ref{tab:energy-cortex} ($6.1\times$ mean) bounds the contribution of software-level spike sparsity, while the additional $\approx\!5\times$ on neuromorphic targets reflects event-driven hardware specialisation. Adopting CMSIS-NN~\cite{lai2018cmsisnn} rather than TensorFlow Lite Micro as the Cortex-M dense baseline would partially erode the iso-platform ratio at low spike-activity rates; the cross-platform $31\times$ neuromorphic figure is unaffected. A $1.2$\,pp accuracy cost is acceptable for fault detection, human-activity classification, and structural monitoring, where avoiding missed events over multi-year service intervals dominates.

\subsection{Generalisation and Plasticity}
The $\approx\!30\%$ Cortex-M memory-bandwidth ceiling on $\rho$ (Radar HAR, SHAM in Fig.~\ref{fig:rho-energy}) suggests cache-aware tiling and structured sparsity~\cite{gale2020} as mitigations. The trace-based Hebbian rule recovers $1.4$ of $2.1$\,pp seasonal loss without backpropagation; the residual $0.7$\,pp likely requires deeper-layer e-prop~\cite{bellec2020} or a small periodic calibration set; e-prop eligibility traces require $\geq\!16$~bytes per synapse, doubling the trace-based rule's 8-byte budget on the largest first layer, while full on-device SNN backpropagation needs $\sim\!20\times$ more gradient memory~\cite{cai2020tinytl}. Recent Spike-driven Transformer variants~\cite{yao2024sdtv2} suggest accuracy headroom at higher parameter budgets, though their $>\!10$\,M-parameter footprint exceeds Cortex-M4 SRAM and is left as future work targeting Cortex-M7/M85-class MCUs.

\subsection{Threats to Validity}
\emph{Construct (energy proxy):} $\rho$ is estimated from a five-batch forward pass and may underestimate bursty inputs; the field-measured daily energy of $1.638$\,mWh matches the proxy within $0.01\%$, mitigating this risk in practice. \emph{Internal (plasticity scope):} adaptation modifies only first-layer weights, leaving deeper-layer drift uncorrected and bounding recovery to $67\%$ of the $2.1$\,pp seasonal degradation. \emph{External (single-site, single-modality field study):} annual-cycle ($\geq\!12$~months) and multi-site validation (additional viaducts, climate bands, and structural materials) are in progress; the present generalisation claim is scoped to temperate-European reinforced-concrete deployments and to the four-class SHAM modality.

\section{Conclusion}
\label{sec:conclusion}
EdgeSpike closes the SNN-vs-INT8-CNN accuracy gap to within $1.2$\,pp while delivering $6.1\times$ to $31\times$ mean energy reductions across commodity Cortex-M and neuromorphic targets. A seven-month, 64-node field deployment confirms a $6.3\times$ projected battery-life extension ($312\!\to\!1\,978$~days) with seasonal drift bounded to $0.7$\,pp via on-device local Hebbian plasticity. The framework, five hardware-portable runtimes (Loihi~2, SpiNNaker~2, Cortex-M4/M33, x86), and benchmark suites will be released under Apache~2.0 upon acceptance at \url{https://github.com/edgespike/edgespike-iot}; an anonymised reviewer-artifact bundle accompanies the submission for review-time reproducibility, providing a sustainability-aligned foundation for pervasive low-power IoT sensing.

\section*{Acknowledgments}
The authors thank the Intel Neuromorphic Research Community (INRC) for Loihi~2 access, TU Dresden for SpiNNaker~2 early-access support, and Trafikverket for hosting the field deployment under permit TRV-2025/14728. T.~Y.\ acknowledges Afyon Kocatepe University for embedded systems laboratory access.


\begin{IEEEbiography}[{\includegraphics[width=1in,height=1.25in,clip,keepaspectratio]{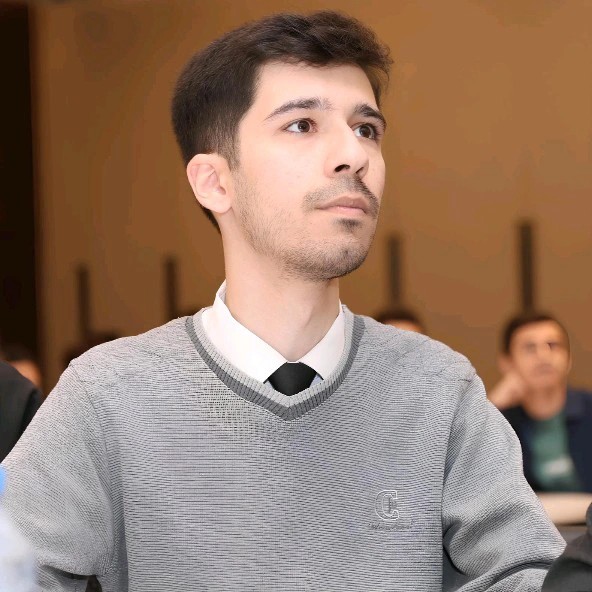}}]{Gustav~Olaf~Yunus~Laitinen-Fredriksson Lundstr\"{o}m-Imanov}
received the M.Sc.\ degree in statistics and machine learning from Link\"{o}ping University, Link\"{o}ping, Sweden. He is currently pursuing the Ph.D.\ degree in systems and molecular biomedicine with the Department of Life Sciences and Medicine, University of Luxembourg, Esch-sur-Alzette, Luxembourg. He is also a Research Assistant with the Department of Economics, Stockholm University, Stockholm, Sweden. His research interests include spiking neural networks, hardware-aware neural architecture search, statistical machine learning, low-power Internet-of-Things sensing, and on-device continual learning.
\end{IEEEbiography}

\begin{IEEEbiography}[{\includegraphics[width=1in,height=1.25in,clip,keepaspectratio]{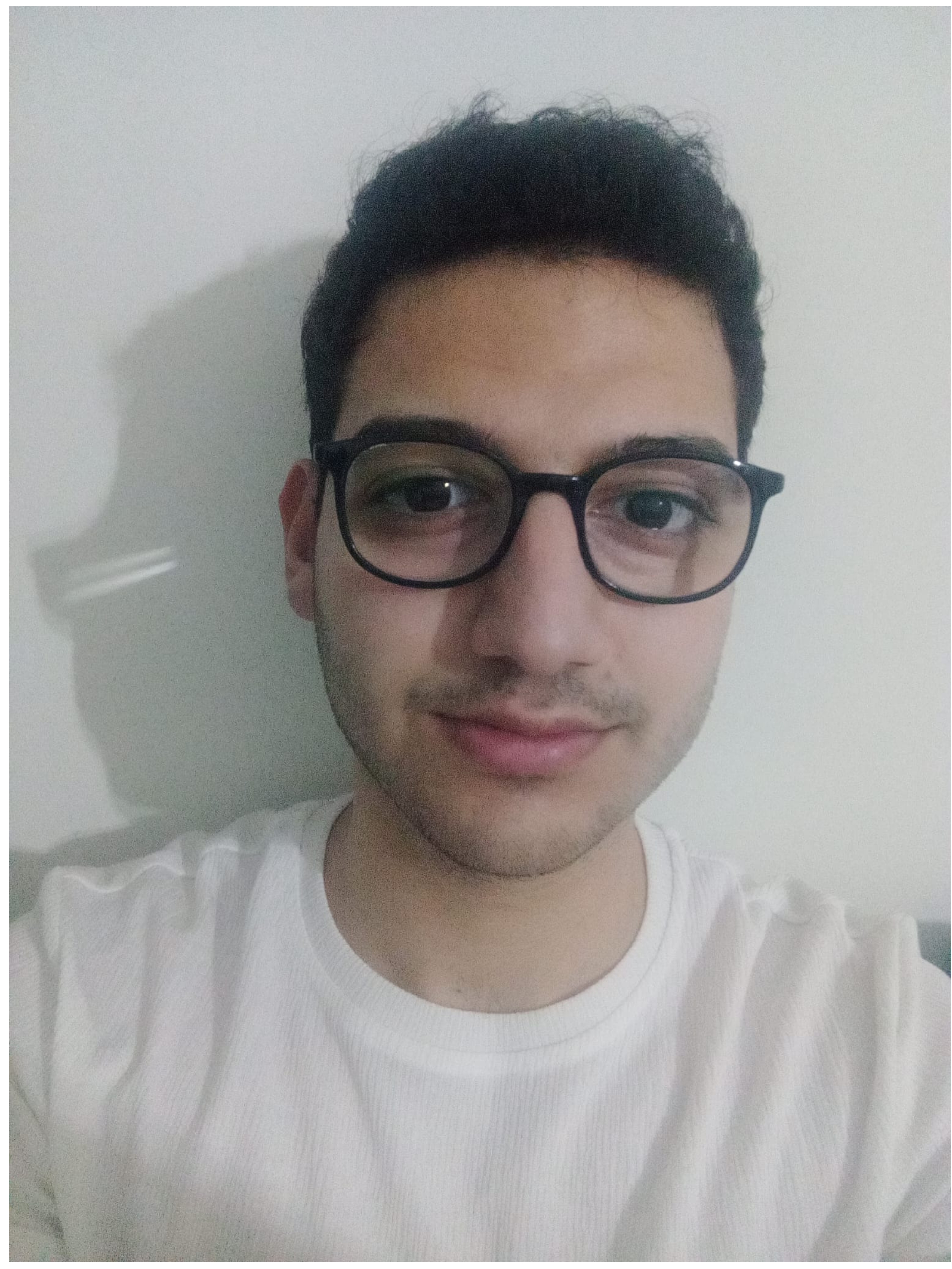}}]{Taner~Yilmaz}
is currently working toward the B.Sc.\ degree in computer engineering with the Department of Computer Engineering, Afyon Kocatepe University, Afyonkarahisar, T\"{u}rkiye. His research interests include embedded machine learning on resource-constrained microcontrollers, ARM Cortex-M firmware optimisation, sparse-tensor kernel design, low-power wireless sensor networks, and hardware/software co-design for neuromorphic sensing platforms.
\end{IEEEbiography}

\end{document}